\documentstyle[aclap]{article}
\newcommand{\hidden}[1]{}
\newcounter{examplectr}
\newcounter{subexamplectr}
\newenvironment{ex}%
   {\vspace{.1in}\addtocounter{examplectr}{1}
     \setcounter{subexamplectr}{0}
     \begin{list}
       {(\arabic{examplectr})}%
       {\setlength{\topsep}{0in}
        \setlength{\leftmargin}{.25in}
               \setlength{\labelsep}{0.075in}}
       \item \begin{minipage}[t]{2.8in} 
   }%
   {\end{minipage}
    \end{list}\vspace{.1in}}
\newenvironment{subex}%
   { \addtocounter{subexamplectr}{1}
     \begin{list}
       {\alph{subexamplectr}}%
       {\setlength{\topsep}{-\parskip}
        \setlength{\leftmargin}{0.175in}
        \setlength{\labelsep}{0.075in}}
       \item
   }%
   {\end{list}}
\newcommand{\exnum}[2]{\addtocounter{examplectr}{#1}(\arabic{examplectr}{#2})\addtocounter{examplectr}{-#1}}
\newcommand{\etal}{{\it et al.}}
\newcommand{\ssp}{\setlength{\baselineskip}{9.95pt}}
\author{Ted Briscoe \\
Computer Laboratory \\ University of Cambridge \\
Pembroke Street, Cambridge CB2 3QG, UK \\ {\tt ejb@cl.cam.ac.uk}\And
John Carroll \\
Cognitive and Computing Sciences \\
University of Sussex \\
Brighton BN1 9QH, UK \\ {\tt john.carroll@cogs.susx.ac.uk}}
\title{\vspace{-0.5in}Automatic Extraction of Subcategorization from Corpora}
\begin{document}
\maketitle
\vspace{-0.5in}
\begin{abstract}
  We describe a novel technique and implemented system for
  constructing a subcategorization dictionary from textual corpora.
  Each dictionary entry encodes the relative frequency of occurrence
  of a comprehensive set of subcategorization classes for English. An
  initial experiment, on a sample of 14 verbs which exhibit multiple
  complementation patterns, demonstrates that the technique achieves
  accuracy comparable to previous approaches, which are all limited to
  a highly restricted set of subcategorization classes.  We also
  demonstrate that a subcategorization dictionary built with the
  system improves the accuracy of a parser by an appreciable
amount\footnote{This work was supported by UK DTI/SALT
project 41/5808 `Integrated Language Database', CEC Telematics Applications
Programme project LE1-2111 `SPARKLE: Shallow PARsing and Knowledge extraction
for Language Engineering', and by SERC/EPSRC
Advanced Fellowships to both authors. We would like to thank the COMLEX Syntax development
team for allowing us access to pre-release data
(for an early experiment), and for useful feedback.}.
\end{abstract}

\section{Motivation}

Predicate subcategorization is a key component of a lexical entry,
because most, if not all, recent syntactic theories `project'
syntactic structure from the lexicon. Therefore, a wide-coverage
parser utilizing such a lexicalist grammar must have access to an
accurate and comprehensive dictionary encoding (at a minimum) the
number and category of a predicate's arguments and ideally also
information about control with predicative arguments, semantic
selection preferences on arguments, and so forth, to allow the
recovery of the correct predicate-argument structure. If the parser
uses statistical techniques to rank analyses, it is also critical that
the dictionary encode the relative frequency of distinct
subcategorization classes for each predicate.

Several substantial machine-readable subcategorization dictionaries
exist for English, either built largely automatically from
machine-readable versions of conventional learners' dictionaries, or
manually by (computational) linguists (e.g.\ the Alvey NL Tools (ANLT)
dictionary, Boguraev \etal\ (1987); the COMLEX Syntax dictionary,
Grishman \etal\ (1994)). Unfortunately, neither approach can yield a
genuinely accurate or comprehensive computational lexicon, because
both rest ultimately on the manual efforts of lexicographers /
linguists and are, therefore, prone to errors of omission and
commission which are hard or impossible to detect automatically (e.g.\ 
Boguraev \& Briscoe, 1989; see also section 3.1 below for an example).
Furthermore, manual encoding is labour intensive and, therefore, it is
costly to extend it to neologisms, information not currently encoded
(such as relative frequency of different subcategorizations), or other
(sub)languages. These problems are compounded by the fact that
predicate subcategorization is closely associated to lexical sense and
the senses of a word change between corpora, sublanguages and/or
subject domains (Jensen, 1991).

In a recent experiment with a wide-coverage parsing system utilizing a
lexicalist grammatical framework, Briscoe \& Carroll (1993) observed
that half of parse failures on unseen test data were caused by
inaccurate subcategorization information in the ANLT dictionary.  The
close connection between sense and subcategorization and between
subject domain and sense makes it likely that a fully accurate
`static' subcategorization dictionary of a language is unattainable in
any case. Moreover, although Schabes (1992) and others have proposed
`lexicalized' probabilistic grammars to improve the accuracy of parse
ranking, no wide-coverage parser has yet been constructed
incorporating probabilities of different subcategorizations for
individual predicates, because of the problems of accurately
estimating them.

These problems suggest that automatic construction or updating of
subcategorization dictionaries from textual corpora is a more
promising avenue to pursue.  Preliminary experiments acquiring a few
verbal subcategorization classes have been reported by Brent (1991,
1993), Manning (1993), and Ushioda \etal\ (1993). In these
experiments the maximum number of distinct subcategorization classes
recognized is sixteen, and only Ushioda \etal\ attempt to derive
relative subcategorization frequency for individual predicates.

We describe a new system capable of distinguishing 160 verbal
subcategorization classes---a superset of those found in the ANLT and
COMLEX Syntax dictionaries. The classes also incorporate information
about control of predicative arguments and alternations such as
particle movement and extraposition. We report an initial experiment
which demonstrates that this system is capable of acquiring the
subcategorization classes of verbs and the relative frequencies of
these classes with comparable accuracy to the less ambitious extant
systems. We achieve this performance by exploiting a more
sophisticated robust statistical parser which yields complete though
`shallow' parses, a more comprehensive subcategorization class
classifier, and {\it a priori} estimates of the probability of
membership of these classes.  We also describe a small-scale
experiment which demonstrates that subcategorization class frequency
information for individual verbs can be used to improve parsing
accuracy.

\section{Description of the System}

\subsection{Overview}

The system consists of the following six components which are applied
in sequence to sentences containing a specific predicate in order to
retrieve a set of subcategorization classes for that predicate:

\begin{enumerate}
  
\item {\bf A tagger}, a first-order HMM part-of-speech (PoS) and
  punctuation tag disambiguator, is used to assign and rank tags for
  each word and punctuation token in sequences of sentences (Elworthy,
  1994).

\item {\bf A lemmatizer} is used to replace word-tag pairs with
  lemma-tag pairs, where a lemma is the morphological base or
  dictionary headword form appropriate for the word, given the PoS
  assignment made by the tagger. We use an enhanced version of the
  {\sc GATE} project stemmer (Cunningham \etal, 1995).
  
\item {\bf A probabilistic LR parser}, trained on a treebank, returns
  ranked analyses (Briscoe \& Carroll, 1993; Carroll, 1993, 1994),
  using a grammar written in a feature-based unification grammar
  formalism which assigns `shallow' phrase structure analyses to tag
  networks (or `lattices') returned by the tagger (Briscoe \& Carroll,
  1994, 1995; Carroll \& Briscoe, 1996).
  
\item {\bf A patternset extractor} which extracts subcategorization
  patterns, including the syntactic categories and head lemmas of
  constituents, from sentence subanalyses which begin/end at the
  boundaries of (specified) predicates.
  
\item {\bf A pattern classifier} which assigns patterns in patternsets
  to subcategorization classes or rejects patterns as unclassifiable
  on the basis of the feature values of syntactic categories and the
  head lemmas in each pattern.
  
\item {\bf A patternsets evaluator} which evaluates sets of
  patternsets gathered for a (single) predicate, constructing putative
  subcategorization entries and filtering the latter on the basis of
  their reliability and likelihood.
\end{enumerate}

For example, building entries for {\it attribute}, and given that one
of the sentences in our data was \exnum{+1}{a}, the tagger and
lemmatizer return \exnum{+1}{b}.
\begin{ex} 
\begin{subex} 
{\it He attributed his failure, he said, to no$<$blank$>$one buying his
books.} 
\end{subex}
\begin{subex} 
{\it he\_PPHS1 attribute\_VVD his\_APP\$ failure\_NN1 ,\_,
he\_PPHS1 say\_VVD ,\_, to\_II no$<$blank$>$one\_PN buy\_VVG his\_APP\$
book\_NN2}
\end{subex} 
\end{ex} 
\vspace{4mm}
\exnum{0}{b} is parsed successfully by the probabilistic LR parser,
and the ranked analyses are returned. Then the patternset extractor
locates the subanalyses containing {\it attribute} and
constructs a patternset. The highest ranked analysis and pattern for this
example are shown in Figure~\ref{attrib}\footnote{The analysis
shows only category aliases rather than sets of feature-value pairs.
{\tt Ta} represents a text adjunct delimited by commas (Nunberg 1990;
Briscoe \& Carroll, 1994). Tokens in the patternset are indexed by sequential
position in the sentence so that two or more tokens of the same type
can be kept distinct in patterns.}.
\begin{figure*} 
\begin{verbatim} 
(Tp                                                         (1 ((((he:1 PPHS1))
   (V2 (N2 he_PPHS1)                                             (VSUBCAT NP_PP)
      (V1 (V0 attribute_VVD)                                     ((attribute:6 VVD))
         (N2 (DT his_APP$)                                       ((failure:8 NN1))
            (N1                                                  ((PSUBCAT SING)
               (N0 (N0 failure_NN1)                               ((to:9 II))
                  (Ta (Pu ,_,)                                     ((no<blank>one:10 PN))
                     (V2 (N2 he_PPHS1)                             ((buy:11 VVG))))
                        (V1 (V0 say_VVD))) (Pu ,_,)))))         . 1))
         (P2 
            (P1 (P0 to_II)
               (N2 no<blank>one_PN) 
               (V1 (V0 buy_VVG) (N2 (DT his_APP$) (N1 (N0 book_NN2)))))))))
\end{verbatim}
\caption{Highest-ranked analysis and patternset for \exnum{0}{b}}
\label{attrib}
\end{figure*}
Patterns encode the value of the VSUBCAT feature from the VP rule and
the head lemma(s) of each argument. In the case of PP ({\tt P2})
arguments, the pattern also encodes the value of PSUBCAT from the PP
rule and the head lemma(s) of its complement(s). In the next stage of
processing, patterns are classified, in this case giving the
subcategorization class corresponding to transitive plus PP with
non-finite clausal complement.

The system could be applied to corpus data by first sorting sentences
into groups containing instances of a specified predicate, but we use
a different strategy since it is more efficient to tag, lemmatize and
parse a corpus just once, extracting patternsets for all predicates in
each sentence; then to classify the patterns in all patternsets; and
finally, to sort and recombine patternsets into sets of patternsets,
one set for each distinct predicate containing patternsets of just the
patterns relevant to that predicate.  The tagger, lemmatizer, grammar
and parser have been described elsewhere (see previous references), so
we provide only brief relevant details here, concentrating on the
description of the components of the system that are new: the
extractor, classifier and evaluator.

The grammar consists of 455 phrase structure rule schemata in the
format accepted by the parser (a syntactic variant of a Definite
Clause Grammar with iterative (Kleene) operators). It is `shallow' in
that no atof which thetempt is made to fully analyse unbounded dependencies.
However, the distinction between arguments and adjuncts is expressed,
following X-bar theory (e.g.\ Jackendoff, 1977), by Chomsky-adjunction
to maximal projections of adjuncts (\mbox{XP $\rightarrow$ XP
Adjunct}) as opposed to `government' of arguments (i.e.\ arguments are
sisters within X1 projections; \mbox{X1 $\rightarrow$ X0 Arg1...
ArgN}). Furthermore, all analyses are rooted (in S) so the grammar
assigns global, shallow and often `spurious' analyses to many
sentences.  There are 29 distinct values for VSUBCAT and 10 for
PSUBCAT; these are analysed in patterns along with specific
closed-class head lemmas of arguments, such as {\it it} (dummy
subjects), {\it whether} (wh-complements), and so forth, to classify
patterns as evidence for one of the 160 subcategorization classes.
Each of these classes can be parameterized for specific predicates by,
for example, different prepositions or particles. Currently, the
coverage of this grammar---the proportion of sentences for which at least one
analysis is found---is 79\% when applied to the Susanne corpus
(Sampson, 1995), a 138K word treebanked and balanced subset of the Brown
corpus. Wide coverage is important since information is
acquired only from successful parses. The combined throughput of the
parsing components on a Sun UltraSparc 1/140 is around 50 words per
CPU second.

\subsection{The Extractor, Classifier and Evaluator}

The extractor takes as input the ranked analyses from the
probabilistic parser. It locates the subanalyses around the predicate,
finding the constituents identified as complements inside each
subanalysis, and the subject clause preceding it. Instances of passive
constructions are recognized and treated specially. The extractor
returns the predicate, the VSUBCAT value, and just the heads of the
complements (except in the case of PPs, where it returns the PSUBCAT
value, the preposition head, and the heads of the PP's complements).

The subcategorization classes recognized by the classifier were
obtained by manually merging the classes exemplified in the COMLEX
Syntax and ANLT dictionaries and adding around 30 classes found by
manual inspection of unclassifiable patterns for corpus examples
during development of the system. These consisted of some extra
patterns for phrasal verbs with complex complementation and with
flexible ordering of the preposition/particle, some for
non-passivizable patterns with a surface direct object, and some for
rarer combinations of governed preposition and complementizer
combinations. The classifier filters out as unclassifiable around 15\%
of patterns found by the extractor when run on all the patternsets
extracted from the Susanne corpus. This demonstrates the value of the
classifier as a filter of spurious analyses, as well as providing both
translation between extracted patterns and two existing
subcategorization dictionaries and a definition of the target
subcategorization dictionary.

The evaluator builds entries by taking the patterns for a given
predicate built from successful parses and records the number of
observations of each subcategorization class.  Patterns provide
several types of information which can be used to rank or select
between patterns in the patternset for a given sentence exemplifying
an instance of a predicate, such as the ranking of the parse from
which it was extracted or the proportion of subanalyses supporting a
specific pattern. Currently, we simply select the pattern supported by
the highest ranked parse. However, we are experimenting with
alternative approaches.  The resulting set of putative classes for a
predicate are filtered, following Brent (1993), by hypothesis testing
on binomial frequency data.

Evaluating putative entries on binomial frequency data requires that
we record the total number of patternsets {\it n} for a given
predicate, and the number of these patternsets containing
a pattern supporting an entry for given class {\it m}. These figures
are straightforwardly computed from the output of the classifier; however, we
also require an estimate of the probability that a pattern for class {\it i}
will occur with a verb which is not a member of subcategorization class {\it
i}. Brent proposes estimating these probabilities experimentally on the basis
of the behaviour of the extractor. We estimate this probability more
directly by first extracting the number of verbs which are members of
each class in the ANLT dictionary (with intuitive estimates for the
membership of the novel classes) and converting this to a probability of class
membership by dividing by the total number of verbs in the
dictionary; and secondly, by multiplying the complement of these
probabilities by the probability of a pattern for class {\it i},
defined as the number of patterns for {\it i} extracted from the
Susanne corpus divided by the total number of patterns. So, $p(v~{\mbox -i})$,
the probability
of verb $v$ not of class $i$ occurring with a pattern for class $i$ is:
\[p(v~{\mbox -i}) = (1 -
\frac{|anlt\_verbs\_in\_class\_i|}{|anlt\_verbs|})
\frac{|patterns\_for\_i|}{|patterns|}\] 
The binomial
distribution gives the probability of an event with probability {\it p}
happening exactly {\it m} times out of
{\it n} attempts:
\[P(m,n,p) = \frac{n!}{m!(n-m)!}p^{m}(1-p)^{n-m}\]
The probability of the event happening
{\it m} or more times is:
\[P(m+,n,p) = \sum_{i=m}^n P(i,n,p)\]
Thus {\it P(m,n,p(v~{\mbox -i}))} is the probability that {\it m} or more
occurrences of patterns for {\it i} will occur with a verb which is
not a member of {\it i}, given {\it n} occurrences of that verb. Setting
a threshold of less than or equal to 0.05 yields a 95\%
or better confidence that a high enough proportion of patterns for
{\it i} have been observed for the verb to be
in class {\it i}\footnote{Brent (1993:249--253) provides a detailed
explanation and justification for the use of this measure.}.

\subsection{Discussion}

Our approach to acquiring subcategorization classes is predicated on
the following assumptions:
\begin{itemize} 
\item most sentences will not allow the application of all possible
rules of English complementation;
\item some sentences will be unambiguous even given the indeterminacy
of the grammar\footnote{In fact, 5\% of sentences in Susanne are assigned only
a single analysis by the grammar.};
\item many incorrect analyses will yield patterns which are
unclassifiable, and are thus filtered out;
\item arguments of a specific verb will occur with greater frequency
than adjuncts (in potential argument positions);
\item the patternset generator will incorrectly output patterns for certain
classes more often than others; and
\item even a highest ranked pattern for {\it i} is only a
probabilistic cue for membership of {\it i}, so membership
should only be inferred if there are enough occurrences of patterns
for {\it i} in the data to outweigh the error probability for {\it i}.
\end{itemize}

This simple automated, hybrid linguistic/statistical approach
contrasts with the manual linguistic analysis of the COMLEX Syntax
lexicographers (Meyers \etal, 1994), who propose five criteria and
five heuristics for argument-hood and six criteria and two heuristics
for adjunct-hood, culled mostly from the linguistics literature. Many
of these are not exploitable automatically because they rest on
semantic judgements which cannot (yet) be made automatically: for
example, optional arguments are often `understood' or implied if
missing.  Others are syntactic tests involving diathesis alternation
possibilities (e.g.\ passive, dative movement, Levin (1993)) which
require recognition that the `same' argument, defined usually by
semantic class / thematic role, is occurring across argument
positions. We hope to exploit this information where possible at a
later stage in the development of our approach. However, recognizing
same/similar arguments requires considerable quantities of lexical
data or the ability to back-off to lexical semantic classes. At the
moment, we exploit linguistic information about the syntactic type,
obligatoriness and position of arguments, as well as the set of
possible subcategorization classes, and combine this with statistical
inference based on the probability of class membership and the
frequency and reliability of patterns for classes.

\section{Experimental Evaluation}

\subsection{Lexicon Evaluation -- Method}

In order to test the accuracy of our system (as developed so far) and
to provide empirical feedback for further development, we took the
Susanne, SEC (Taylor \& Knowles, 1988) and LOB corpora (Garside \etal,
1987)---a total of 1.2 million words---and extracted all sentences
containing an occurrence of one of fourteen verbs, up to a maximum of
1000 citations of each. These verbs, listed in Figure~\ref{rawres},
were chosen at random, subject to the constraint that they exhibited
multiple complementation patterns. The sentences containing these
verbs were tagged and parsed automatically, and the extractor,
classifier and evaluator were applied to the resulting successful
analyses. The citations from which entries were derived totaled
approximately 70K words.

The results were evaluated against a merged entry for these verbs from
the ANLT and COMLEX Syntax dictionaries, and also against a manual
analysis of the corpus data for seven of the verbs. The process of
evaluating the performance of the system relative to the dictionaries
could, in principle, be reduced to an automated report of {\it
type precision} (percentage of correct subcategorization classes to all
classes found) and {\it recall} (percentage of correct classes found in the
dictionary entry). However, since there are disagreements between the
dictionaries and there are classes found in the corpus data that are not
contained in either dictionary, we report results relative both to a
manually merged entry from ANLT and COMLEX, and also, for seven of the verbs,
to a manual analysis of the actual corpus data. The latter analysis is
necessary because precision and recall measures against the merged entry
will still tend to yield inaccurate results as the system cannot acquire
classes not exemplified in the data, and may acquire classes incorrectly
absent from the dictionaries.  

We illustrate these problems with reference to {\it seem}, where there
is overlap, but not agreement between the COMLEX and ANLT entries.
Thus, both predict that {\it seem} will occur with a sentential
complement and dummy subject, but only ANLT predicts the possibility
of a `wh' complement and only COMLEX predicts the (optional) presence
of a PP[to] argument with the sentential complement. One ANLT entry
covers two COMLEX entries given the different treatment of the
relevant complements but the classifier keeps them distinct. The
corpus data for {\it seem} contains examples of further classes which
we judge valid, in which {\it seem} can take a PP[to] and infinitive
complement, as in {\it he seems to me to be insane}, and a passive
participle, as in {\it he seemed depressed}. This comparison
illustrates the problem of errors of omission common to computational
lexicons constructed manually and also from machine-readable
dictionaries. All classes for {\it seem} are exemplified in the corpus
data, but for {\it ask}, for example, eight classes (out of a possible
27 in the merged entry) are not present, so comparison only to the
merged entry would give an unreasonably low estimate of recall.

\subsection{Lexicon Evaluation -- Results}

Figure~\ref{rawres} gives the raw results for the merged entries and
corpus analysis on each verb.  It shows the number of {\it true
positives} (TP), correct classes proposed by our system, {\it false
positives} (FP), incorrect classes proposed by our system, and {\it
false negatives} (FN), correct classes not proposed by our system, as
judged against the merged entry, and, for seven of the verbs, against
the corpus analysis.  It also shows, in the final column, the number
of sentences from which classes were extracted.
\begin{figure*}
\begin{center}
\begin{tabular}{||l||r|r|r||r|r|r||r||} \hline 
         & \multicolumn{3}{c||}{Merged Entry} & \multicolumn{3}{c||}{Corpus Data} & \multicolumn{1}{c||}{No. of}\\
         & TP & FP & FN   & TP & FP & FN   & \multicolumn{1}{c||}{Sentences} \\ \hline
ask      &  9 & 0 & 18    &  9 & 0 & 10    &  390\\
begin    &  4 & 1 & 7     &  4 & 1 & 7     &  311\\
believe  &  4 & 4 & 11    &  4 & 4 & 8     &  230\\
cause    &  2 & 3 & 6     &  2 & 3 & 5     &   95\\
expect   &  6 & 5 & 3     & -- & -- & --   &  223\\
find     &  5 & 7 & 15    & -- & -- & --   &  645\\
give     &  5 & 2 & 11    &  5 & 2 & 5     &  639\\
help     &  6 & 3 & 8     & -- & -- & --   &  223\\
like     &  3 & 2 & 7     & -- & -- & --   &  228\\
move     &  4 & 3 & 9     & -- & -- & --   &  217\\
produce  &  2 & 1 & 3     & -- & -- & --   &  152\\
provide  &  3 & 2 & 6     & -- & -- & --   &  217\\
seem     &  8 & 1 & 4     &  8 & 1 & 4     &  534\\
swing    &  4 & 0 & 10    &  4 & 0 & 8     &   45\\ \hline
Totals   & 65 & 34 & 118  & 36 & 11 & 47   & 4149\\ \hline
\end{tabular}
\caption{Raw results for test of 14 verbs}
\label{rawres}
\end{center}
\end{figure*}

Figure~\ref{type-prec-rec} gives the type precision and recall of our
system's recognition of subcategorization classes as evaluated against
the merged dictionary entries (14 verbs) and against the manually
analysed corpus data (7 verbs). The frequency distribution of the
classes is highly skewed: for example for {\it believe}, there are 107
instances of the most common class in the corpus data, but only 6
instances in total of the least common four classes. More generally,
for the manually analysed verbs, almost 60\% of the false negatives
have only one or two exemplars each in the corpus citations. None of
them are returned by the system because the binomial filter always
rejects classes hypothesised on the basis of such little evidence.
\begin{figure}
\begin{center}
\begin{tabular}{||l|r|r||} \hline
           & Dictionary & Corpus\\
           & (14 verbs) & (7 verbs)\\ \hline
Precision  & 65.7\%     & 76.6\%\\
Recall     & 35.5\%     & 43.4\%\\ \hline
\end{tabular}
\caption{Type precision and recall}
\label{type-prec-rec}
\end{center}
\end{figure}

In Figure~\ref{ranking} we estimate the accuracy with which our system
\begin{figure}
\begin{center}
\begin{tabular}{||l|r||} \hline
         & Ranking Accuracy\\ \hline
ask      & 75.0\%\\
begin    & 100.0\%\\
believe  & 66.7\%\\
cause    & 100.0\%\\
give     & 70.0\%\\
seem     & 75.0\%\\
swing    & 83.3\%\\ \hline
Mean     & 81.4\%\\ \hline
\end{tabular}
\caption{Ranking accuracy of classes}
\label{ranking}
\end{center}
\vspace{-0.5cm}
\end{figure}
ranks true positive classes against the correct ranking for the seven
verbs whose corpus input was manually analysed. We compute this
measure by calculating the percentage of pairs of classes at positions
$(n,m)$ s.t.\ $n<m$ in the system ranking that are ordered the same in
the correct ranking. This gives us an estimate of the accuracy of the
relative frequencies of classes output by the system.

For each of the seven verbs for which we undertook a corpus analysis,
we calculate the token recall of our system as the percentage (over
all exemplars) of true positives in the corpus.  This gives us an
estimate of the parsing performance that would result from providing a
parser with entries built using the system, shown in
Figure~\ref{token-rec}.
\begin{figure}
\begin{center}
\begin{tabular}{||l|r||} \hline
         & Token Recall\\ \hline
ask      & 78.5\%\\
begin    & 73.8\%\\
believe  & 34.5\%\\
cause    & 92.1\%\\
give     & 92.2\%\\
seem     & 84.7\%\\
swing    & 39.2\%\\ \hline
Mean     & 80.9\%\\ \hline
\end{tabular}
\caption{Token recall}
\label{token-rec}
\end{center}
\vspace{-0.4cm}
\end{figure}

Further evaluation of the results for these seven verbs reveals that
the filtering phase is the weak link in the system. There are only 13
{\it true negatives} which the system failed to propose,
each exemplified in the data by a mean of 4.5 examples. On the other hand,
there are 67 {\it false negatives} supported by an estimated mean of
7.1 examples which should, ideally, have been accepted by the filter,
and 11 {\it false positives} which should have been rejected. The
performance of the filter for classes with less than 10 exemplars is
around chance, and a simple heuristic of accepting all classes with
more than 10 exemplars would have produced broadly similar results
for these verbs. The filter may well be performing poorly because the
probability of generating a subcategorization class for a given verb
is often lower than the error probability for that class.

\subsection{Parsing Evaluation}

In addition to evaluating the acquired subcategorization information
against existing lexical resources, we have also evaluated the
information in the context of an actual parsing system. In particular
we wanted to establish whether the subcategorization frequency
information for individual verbs could be used to improve the accuracy
of a parser that uses statistical techniques to rank analyses.

The experiment used the same probabilistic parser and tag sequence
grammar as are present in the acquisition system (see references
above)---although the experiment does not in any way rely on the
parsers or grammars being the same. We randomly selected a test set of
250 in-coverage sentences (of lengths 3--56 tokens, mean 18.2) from
the Susanne treebank, retagged with possibly multiple tags per word,
and measured the `baseline' accuracy of the unlexicalized parser on
the sentences using the now standard PARSEVAL/GEIG evaluation metrics
of mean crossing brackets per sentence and (unlabelled) bracket recall
and precision (e.g.\ Grishman \etal, 1992); see
figure~\ref{parse-eval}\footnote{Carroll \& Briscoe (1996) use the
  same test set, although the baseline results reported here differ
  slightly due to differences in the mapping from parse trees to
  Susanne-compatible bracketings.}.
\begin{figure}
\centering
\begin{tabular}{||l|rrr||} \hline
       & \multicolumn{1}{c}{Mean} &
\multicolumn{1}{c}{Recall} & \multicolumn{1}{c||}{Precision} \\

       & \multicolumn{1}{c}{crossings} &&
\\ \hline

`Baseline'
       & 1.00          & 70.7\%         & 72.3\% \\
Lexicalised
       & 0.93          & 71.4\%         & 72.9\% \\ \hline
\end{tabular}
\caption{GEIG evaluation metrics for parser against Susanne bracketings}
\label{parse-eval}
\end{figure}
Next, we collected all words in the test corpus tagged as possibly being
verbs (giving a total of 356 distinct lemmas) and retrieved all citations of
them in the LOB corpus, plus Susanne with the 250 test sentences excluded. We
acquired subcategorization and associated frequency information from the
citations, in the process successfully parsing 380K words. We then parsed
the test set, with each verb subcategorization possibility weighted by its
raw frequency score, and using the naive add-one smoothing technique to allow
for omitted possibilities. The GEIG measures for the lexicalized parser show
a 7\% improvement in the crossing bracket score
(figure~\ref{parse-eval}).  Over the existing test corpus this is not
statistically significant at the 95\% level ({\it paired~t-test}, 1.21,
249~$df$, $p=0.11$)---although if the pattern of differences were
maintained over a larger test set of 470 sentences it would be
significant. We expect that a more sophisticated smoothing technique, a
larger acquisition corpus, and extensions to the system to deal with
nominal and adjectival predicates would improve accuracy still
further. Nevertheless, this experiment demonstrates that lexicalizing a
grammar/parser with subcategorization frequencies can appreciably improve
the accuracy of parse ranking.

\section{Related Work}

Brent's (1993) approach to acquiring subcategorization is based on a
philosophy of only exploiting unambiguous and determinate information
in unanalysed corpora. He defines a number of lexical patterns (mostly
involving closed class items, such as pronouns) which reliably cue one
of five subcategorization classes. Brent does not report
comprehensive results, but for one class, sentential complement verbs,
he achieves 96\% precision and 76\% recall at classifying individual
tokens of 63 distinct verbs as exemplars or non-exemplars of this
class. He does not attempt to rank different classes for a given verb.

Ushioda \etal\ (1993) utilise a PoS tagged corpus and
finite-state NP parser to recognize and calculate the relative
frequency of six subcategorization classes. They report an accuracy
rate of 83\% (254 errors) at classifying 1565 classifiable tokens of
33 distinct verbs in running text and suggest that incorrect noun
phrase boundary detection accounts for the majority of errors. They
report that for 32 verbs their system correctly predicts the most
frequent class, and for 30 verbs it correctly predicts the second most
frequent class, if there was one. Our system rankings include all
classes for each verb, from a total of 160 classes, and average 81.4\%
correct.

Manning (1993) conducts a larger experiment, also using a PoS tagged
corpus and a finite-state NP parser, attempting to recognize sixteen
distinct complementation patterns.  He reports that for a test
sample of 200 tokens of 40 verbs in running text, the acquired
subcategorization dictionary listed the appropriate entry for 163
cases, giving a token recall of 82\% (as compared with 80.9\% in our
experiment). He also reports a comparison of acquired entries for
the verbs to the entries given in the {\it Oxford Advanced Learner's
Dictionary of Current English} (Hornby, 1989) on which his
system achieves a precision of 90\% and a recall of 43\%. His system
averages 3.48 subentries (maximum 10)---less then half the number
produced in our experiment. It is not clear what level of evidence the
performance of Manning's system is based on, but the system was
applied to 4.1 million words of text (c.f.\ our 1.2 million words) and
the verbs are all common, so it is likely that considerably more
exemplars of each verb were available.

\section{Conclusions and Further Work}

The experiment and comparison reported above suggests that our more
comprehensive subcategorization class extractor is able both to assign
classes to individual verbal predicates and also to rank them according to
relative frequency with comparable accuracy to extant systems. We have
also demonstrated that a subcategorization dictionary built with the
system can improve the accuracy of a probabilistic parser by an
appreciable amount.

The system we have developed is straightforwardly extensible to
nominal and adjectival predicates; the existing grammar distinguishes
nominal and adjectival arguments from adjuncts structurally, so all
that is required is extension of the classifier. Developing an
analogous system for another language would be harder but not
infeasible; similar taggers and parsers have been developed for a
number of languages, but no extant subcategorization dictionaries
exist to our knowledge, therefore the lexical statistics we utilize for
statistical filtering would need to be estimated, perhaps using the
technique described by Brent (1993). However, the entire approach to
filtering needs improvement, as evaluation of our results demonstrates
that it is the weakest link in our current system.

Our system needs further refinement to narrow some subcategorization
classes, for example, to choose between differing control options with
predicative complements. It also needs supplementing with information
about diathesis alternation possibilities (e.g.\ Levin, 1993) and
semantic selection preferences on argument heads. Grishman \&
Sterling (1992), Poznanski \& Sanfilippo (1993), Resnik (1993), Ribas
(1994) and others have shown that it is possible to acquire selection
preferences from (partially) parsed data. Our system already gathers
head lemmas in patterns, so any of these approaches could be applied,
in principle.  In future work, we intend to extend the system in this
direction. The ability to recognize that argument slots of different
subcategorization classes for the same predicate share semantic
restrictions/preferences would assist recognition that the predicate
undergoes specific alternations, this in turn assisting
inferences about control, equi and raising (e.g.\ Boguraev \& Briscoe,
1987).

\section*{References}
\newcommand{\book}[4]{\item #1 #4. {\it #2}. #3.}
\newcommand{\barticle}[7]{\item #1 #7. #2. In #5 eds. {\it #4}. #6:~#3.}
\newcommand{\bparticle}[6]{\item #1 #6. #2. In #4 eds. {\it #3}. #5.}
\newcommand{\boarticle}[5]{\item #1 #5. #2. In {\it #3}. #4.}
\newcommand{\farticle}[6]{\item #1 #6. #2. In #4 eds. {\it #3}:~#5. Forthcoming.}
\newcommand{\uarticle}[5]{\item #1 #5. #2. In #4 eds. {\it #3}. Forthcoming.}
\newcommand{\jarticle}[6]{\item #1 #6. #2. {\it #3} #4:~#5.}
\newcommand{\particle}[6]{\item #1 #6. #2. In {\it Proceedings of the
#3},~#4. #5.}
\newcommand{\lazyparticle}[5]{\item #1 #5. #2. In {\it Proceedings of the
#3}, #4.}
\newcommand{\lazyjarticle}[4]{\item #1 #4. #2. {\it #3}.}
\newcommand{\lazyfjarticle}[4]{\item #1 #4. #2. {\it #3}. Forthcoming.}
\newcommand{\bookartnopp}[6]{\item #1 (#6) #2 In #4 (ed.), {\it #3,} #5.}

\ssp

\begin{list}{}
   {\leftmargin 0pt
    \itemindent 0pt
    \itemsep 2pt plus 1pt
    \parsep 2pt plus 1pt}

\jarticle{Boguraev, B. \& Briscoe, E.}
         {Large lexicons for natural language processing:
utilising the grammar coding system of the {\it Longman Dictionary of Contemporary English}}
         {Computational Linguistics} 
         {13.4}
         {219--240} 
         {1987}

\barticle{Boguraev, B. \& Briscoe, E.}
         {Introduction}
         {1--40}
         {Computational Lexicography for Natural Language Processing}
         {Boguraev, B. \& Briscoe, E.}
         {Longman, London}
         {1989}

\particle{Boguraev, B., Briscoe, E., Carroll, J., Carter, D. \&
Grover, C.}
{The derivation of a gram-matically-indexed lexicon from the Longman Dictionary of
Contemporary English}
{25th Annual Meeting of the Association for Computational Linguistics}
{Stanford, CA}
{193--200}
{1987}

\particle{Brent, M.}
         {Automatic acquisition of subcategorization frames from untagged text}
         {29th Annual Meeting of the Association for Computational Linguistics}
         {Berkeley, CA}
         {209--214}
         {1991}

\jarticle{Brent, M.}
         {From grammar to lexicon: unsupervised learning of lexical syntax}
         {Computational Linguistics}
         {19.3}
         {243--262}
         {1993}

\jarticle{Briscoe, E. \& Carroll, J.}
     {Generalised probabilistic LR parsing for unification-based grammars}
     {Computational Linguistics}
     {19.1}
     {25--60}
     {1993}

\book{Briscoe, E. \& Carroll, J.}
     {Parsing (with) punctuation}
     {Rank Xerox Research Centre, Grenoble, MLTT-TR-007}
     {1994}

\particle{Briscoe, E. \& Carroll, J.}
{Developing and evaluating a probabilistic LR parser of part-of-speech
and punctuation labels}
{4th ACL/SIGPARSE International Workshop on Parsing Technologies}
{Prague, Czech Republic}
{48--58}
{1995}

\book{Carroll, J.}
     {Practical unification-based parsing of natural language}
     {Cambridge University Computer Laboratory, TR-224}
     {1993}

\particle{Carroll, J.}
{Relating complexity to practical performance in parsing with wide-coverage unification grammars}
{32nd Annual Meeting of the Association for Computational Linguistics}
{NMSU, Las Cruces, NM}
{287--294}
{1994}

\particle{Carroll, J. \& Briscoe, E.}
         {Apportioning development effort in a probabilistic LR parsing
system through evaluation}
         {ACL SIGDAT Conference on Empirical Methods in Natural Language
Processing}
         {University of Pensylvania, Philadelphia, PA}
         {92--100}
         {1996}

\barticle{Carroll, J. \& Grover, C.}
         {The derivation of a large computational lexicon for English 
          from LDOCE} 
         {117--134}
         {Computational Lexicography for Natural Language Processing}
         {Boguraev, B. and Briscoe, E.}
         {Longman, London}
         {1989}

\book{Cunningham, H., Gaizauskas, R. \& Wilks, Y.}
{A general architecture for text engineering (GATE) - a new approach to
language R\&D}
{Research memo CS-95-21, Department of Computer Science, University of
Sheffield, UK}
{1995}

\particle{de Marcken, C.}
{Parsing the LOB corpus}
{28th Annual Meeting of the Association for Computational Linguistics}
{Pittsburgh, PA}
{243--251}
{1990}
      
\lazyparticle{Elworthy, D.}
         {Does Baum-Welch re-estimation help taggers?}
         {4th Conf. Applied NLP}
         {Stuttgart, Germany}
         {1994}

\book{Garside, R., Leech, G. \& Sampson, G.}
     {The computational analysis of English: A corpus-based approach}
     {Longman, London}
     {1987}

\particle{Grishman, R., Macleod, C. \& Meyers, A.}
         {Comlex syntax: building a computational lexicon}
         {International Conference on Computational Linguistics, COLING-94}
         {Kyoto, Japan}
         {268--272}
         {1994}

\particle{Grishman, R., Macleod, C. \& Sterling, J.}
{Evaluating parsing strategies using standardized parse files}
{3rd ACL Conference on Applied Natural Language Processing}
{Trento, Italy}
{156--161}
{1992}

\particle{Grishman, R. \& Sterling, J.}
         {Acquisition of selectional patterns}
         {International Conference on Computational Linguistics, COLING-92}
         {Nantes, France}
         {658--664}
         {1992}

\book{Jackendoff, R.}
     {X-bar syntax}
     {MIT Press; Cambridge, MA.}
     {1977}

\bparticle{Jensen, K.}
         {A broad-coverage natural language analysis system}
         {Current Issues in Parsing Technology}
         {M. Tomita}
         {Kluwer, Dordrecht}
         {1991}

\book{Levin,  B.}
{Towards a lexical organization of English verbs}
{Chicago University Press, Chicago}
{1993}

\particle{Manning, C.}
         {Automatic acquisition of a large subcategorisation
dictionary from corpora}
         {31st Annual Meeting of the Association for Computational Linguistics}
         {Columbus, Ohio}
         {235--242}
         {1993}

\book{Meyers, A., Macleod, C. \& Grishman, R.}
     {Standardization of the complement adjunct distinction}
     {New York University, Ms}
     {1994}

\book{Nunberg, G.}
     {The linguistics of punctuation}
     {CSLI Lecture Notes 18, Stanford, CA}
     {1990}

\lazyparticle{Poznanski, V. \& Sanfilippo, A.}
         {Detecting dependencies between semantic verb subclasses and
subcategorization frames in text corpora}
         {SIGLEX ACL Workshop on the Acquisition of Lexical Knowledge from Text}
         {Boguraev, B. \& Pustejovsky, J. eds}
         {1993}

\book{Resnik, P.}
     {Selection and information: a class-based approach to lexical 
relationships}
     {University of Pennsylvania, CIS Dept, PhD thesis}
     {1993}

\lazyparticle{Ribas, P.}
         {An experiment on learning appropriate selection restrictions from
a parsed corpus}
         {International Conference on Computational Linguistics, COLING-94}
         {Kyoto, Japan}
         {1994}

\book{Sampson, G.} {English for the computer} {Oxford, UK: Oxford University
Press} {1995}

\particle{Schabes, Y.}
         {Stochastic lexicalized tree adjoining grammars}
         {International Conference on Computational Linguistics, COLING-92}
         {Nantes, France}
         {426--432}
         {1992}
 
\book{Taylor, L. \& Knowles, G.}
     {Manual of information to accompany the SEC corpus:
the machine-readable corpus of spoken English}
     {University of Lancaster, UK, Ms}
     {1988}

\barticle{Ushioda, A., Evans, D., Gibson, T. \& Waibel, A.}
         {The automatic acquisition of frequencies of verb
subcategorization frames from tagged corpora}
         {95--106}
         {SIGLEX ACL Workshop on the Acquisition of Lexical Knowledge from Text}
         {Boguraev, B. \& Pustejovsky, J.}
         {Columbus, Ohio}
         {1993}

\end{list}
\end{document}